\documentclass[letterpaper, 10 pt, conference]{ieeeconf}  %

\IEEEoverridecommandlockouts                              %

\usepackage{graphicx}
\usepackage{amsmath}
\usepackage{amssymb}
\usepackage{booktabs}
\usepackage{microtype}

\usepackage[capitalize]{cleveref}
\crefname{section}{Sec.}{Secs.}
\Crefname{section}{Section}{Sections}
\Crefname{table}{Table}{Tables}
\crefname{table}{Tab.}{Tabs.}

\newlength\savewidth\newcommand\shline{\noalign{\global\savewidth\arrayrulewidth
  \global\arrayrulewidth 1pt}\hline\noalign{\global\arrayrulewidth\savewidth}}

\newcommand{\eg}{\textit{e}.\textit{g}.}
\newcommand{\ie}{\textit{i}.\textit{e}.}

\title{\LARGE \bf
Superpixel Transformers for Efficient Semantic Segmentation
}

\author{Alex Zihao Zhu$^{1*}$, Jieru Mei$^{2*}$, Siyuan Qiao$^{3}$, Hang Yan$^{1}$,\\
Yukun Zhu$^{3}$, Liang-Chieh Chen$^{4}$, Henrik Kretzschmar$^{5}$%
\thanks{$^{1}$Waymo LLC}%
\thanks{$^{2}$Johns Hopkins University (Work done as an intern at Waymo)}%
\thanks{$^{3}$Google Research}%
\thanks{$^{4}$ByteDance Research (Work done while at Google Research)}%
\thanks{$^{5}$Work done while at Waymo}
\thanks{$^*$Equal contributions}%
}

\begin{document}

\maketitle
\thispagestyle{empty}
\pagestyle{empty}

\begin{abstract}
Semantic segmentation, which aims to classify every pixel in an image, is a key task in machine perception, with many applications across robotics and autonomous driving.
Due to the high dimensionality of this task, most existing approaches use local operations, such as convolutions, to generate per-pixel features.
However, these methods are typically unable to effectively leverage global context information due to the high computational costs of operating on a dense image.
In this work, we propose a solution to this issue by leveraging the idea of superpixels, an over-segmentation of the image, and applying them with a modern transformer framework. 
In particular, our model learns to decompose the pixel space into a spatially low dimensional superpixel space via a series of local cross-attentions.
We then apply multi-head self-attention to the superpixels to enrich the superpixel features with global context and then directly produce a class prediction for each superpixel.
Finally, we directly project the superpixel class predictions back into the pixel space using the associations between the superpixels and the image pixel features.
Reasoning in the superpixel space allows our method to be substantially more computationally efficient compared to convolution-based decoder methods. Yet, our method achieves state-of-the-art performance in semantic segmentation due to the rich superpixel features generated by the global self-attention mechanism.
Our experiments on Cityscapes and ADE20K demonstrate that our method matches the state of the art in terms of accuracy, while outperforming in terms of model parameters and latency.

\end{abstract}
\section{Introduction}
The problem of semantic segmentation, or classifying every pixel in the image, is increasingly common in many robotics applications. A dense, fine-grained, understanding of the world is necessary for navigation in cluttered environments, particularly for applications such as autonomous driving, where scene understanding is deeply safety-critical. On the other hand, many robotics systems are combinations of highly complex and specialized systems, and latency is an ever-present issue for real time operation.

The balance between safety and performance and latency is critical for modern robotic systems. While the state of the art in semantic segmentation is able to achieve strong performance in terms of metrics such as mean Intersection over Union (mIoU), many methods still rely on dense decoders which produce predictions for every pixel in the scene. As a result, these methods tend to be relatively expensive, and arguably produce a lot of redundant computation for nearby pixels that are often very similar.

To address this issue, we aim to bring the classical ideas surrounding superpixels into modern deep learning. The premise of using superpixels is to decompose and over-segment the image into a series of irregular patches. By grouping similar pixels into superpixels and then operating on the superpixel level, one can significantly reduce the computational cost of dense prediction tasks, such as semantic segmentation. Classical superpixel algorithms, such as SLIC~\cite{achanta2012slic}, however, rely on hard associations between each image pixel and superpixel. This makes it hard to embed the superpixel representation into neural network architectures~\cite{lecun1998gradient} as this association is not differentiable for back propagation. Recent works, such as superpixel sampling networks (SSN)~\cite{jampani2018superpixel}, resolve this issue by turning the hard association into a soft one.
While this is a step towards incorporating superpixels into neural networks, their segmentation quality still lags behind other models that adopt the per-pixel or per-mask representation.
\begin{figure}[t!]
  \begin{center}
  \includegraphics[width=\linewidth]{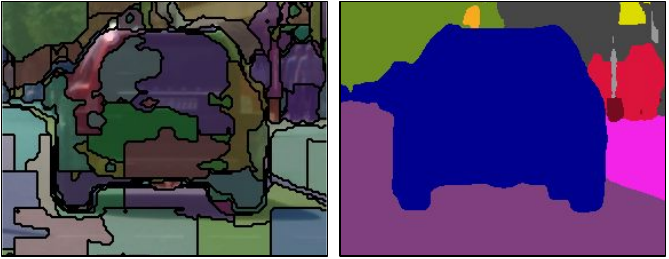} 
  \end{center}
  \caption{Our method enables efficient segmentation of high-resolution camera images by learning to decompose the images into a set of superpixels. Specifically, we oversegment the image pixels into a small set of soft superpixels (left) via a series of local cross attentions. The superpixels are then refined via a set of multi-head self attentions, and directly classified. Finally, we fuse the class predictions with the superpixel-pixel associations to produce a dense semantic segmentation (right). %
  }  
  \label{fig:teaser}   
\end{figure}
\begin{figure*}[t!]
  \begin{center}
  \includegraphics[width=\linewidth]{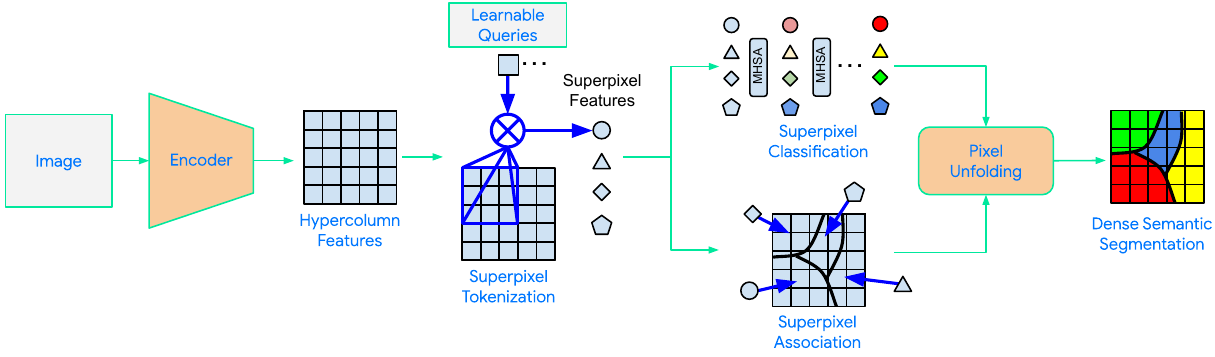} 
  \end{center}
  \caption{Our proposed Superpixel Transformer architecture. Given an image, we first generate hypercolumn features with an off-the-shelf encoder backbone. Our superpixel tokenization module uses a series of local dual-path cross-attentions to generate features for each superpixel.
  The superpixel features are then enriched by several multi-head self-attention (MHSA) layers to produce a class prediction for each superpixel, while the associations between each superpixel and pixel feature are computed from their respective features.
  Finally, the superpixel class predictions are unfolded into the dense pixel space using the associations.
  Note that the figure illustrates a hard assignment between pixels and superpixels for simplicity, while in practice we apply a differentiable soft assignment.}  
  \label{fig:architecture}   
\end{figure*}

In this work, we propose a novel architecture that aims to revive the differentiable superpixel generation pipeline in a modern transformer framework~\cite{vaswani2017attention}. In place of the iterative clustering algorithms used in SLIC~\cite{achanta2012slic} and SSN~\cite{jampani2018superpixel},
we propose to learn the superpixel representation by developing a series of \textit{local} cross-attentions between a set of learned superpixel queries and pixel features.
The outputs of cross-attention modules act as the superpixel features, directly used for semantic segmentation prediction.
As a result, the proposed transformer decoder effectively converts object queries to superpixel features, enabling the model to learn the superpixel representation end-to-end.

Operating on the superpixel level provides a number of notable benefits. Conventional pixel-based approaches are limited by the high dimensionality of the pixel space, making global self-attention computationally intractable. Numerous approaches such as axial attention~\cite{wang2020axial} or window attention~\cite{liu2021swin} have been developed to work around these issues by relaxing the global attention to a local one. By over-segmenting the image into a small set of superpixels, we are able to efficiently apply \textit{global} self-attention on the superpixels, providing full global context to the superpixel features, even when reasoning about high-resolution images. Despite applying global self-attention (vs conventional convolutional neural networks) in our model, our method is more efficient than existing methods due to the low dimensionality of the superpixel space. Finally, we directly produce semantic classes for the superpixel features, and then back-project the predicted classes onto the image space using the superpixel-pixel associations.

We perform extensive evaluations on the Cityscapes~\cite{cordts2016cityscapes} and ADE20K~\cite{zhou2017scene} datasets, where our method matches state-of-the-art performance, but at significantly lower computational cost.

In summary, the main contributions of this work are as follows:
\begin{itemize}
    \item The first work that revives the superpixel representation in the modern transformer framework, where the object queries are used to learn superpixel features.
    \item A novel network architecture that uses local cross-attention to significantly reduce the spatial dimensionality of pixel features to a small set of superpixel features, enabling learning the global context between them and the direct classification of each superpixel.
    \item A superpixel association and unfolding scheme that projects each superpixel class prediction back to a dense pixel segmentation, discarding the CNN pixel decoder.
    \item Experiments on the Cityscapes and ADE20k datasets, where our method outperforms the state of the art at substantially lower computational cost.
\end{itemize}
\section{Related Work}
\textit{Superpixels for Segmentation}~\quad
Before the deep learning era, the superpixel representation, paired with graphical models, was the main paradigm for image segmentation.
Superpixel methods~\cite{lloyd1982least,shi2000normalized, comaniciu2002mean, achanta2012slic} are usually used in the pre-processing step to reduce the computation cost.
A shallow classifier, \begin{it}e.g.\end{it}, SVM~\cite{cortes1995support}, predicts the semantic labels of each superpixel~\cite{fulkerson2009class}, which
aggregates hand-crafted features.
The graphical models, particularly conditional random fields~\cite{lafferty2001conditional}, are then employed to refine the segmentation results~\cite{he2004multiscale,ladicky2009associative,krahenbuhl2011efficient}.

\textit{ConvNets for Segmentation}~\quad
Convolutional neural networks (ConvNets)~\cite{lecun1998gradient} deployed in a fully convolutional manner~\cite{long2014fully} performs semantic segmentation by pixel-wise classification.
Typical ConvNet-based approaches include the DeepLab series~\cite{deeplabv12015,chen2017deeplabv3,deeplabv3plus2018}, PSPNet~\cite{zhao2017pyramid}, UPerNet~\cite{xiao2018unified}, and OCRNet~\cite{yuan2020object}.
Alternatively, there are some works~\cite{gadde2016superpixel,jampani2018superpixel} that employ superpixels to aggregate features extracted by ConvNets and show promising results.

\textit{Transformers for Segmentation}~\quad
Transformers~\cite{vaswani2017attention} and their vision variants~\cite{dosovitskiy2020image} have been adopted as the backbone encoders for image segmentation~\cite{chen2016attention,zheng2021rethinking,chen2021transunet}.
Transformer encoders can be instantiated as augmenting ConvNets with self-attention modules~\cite{wang2018non,wang2020axial}.
When used as stand-alone backbones~\cite{ramachandran2019stand,dosovitskiy2020image,liu2021swin,xie2021segformer}, they also demonstrate strong performance compared to the previous ConvNet baselines.

Transformers are also used as the decoders~\cite{carion2020end} for image segmentation.
A popular design is to generate masks embedding vectors from object queries and then multiply them with the pixel features to generate masks~\cite{tian2020conditional,wang2020solov2}.
For example, MaX-DeepLab~\cite{wang2021max} proposes an end-to-end mask transformer framework that directly predicts class-labeled object masks.
Segmenter~\cite{strudel2021segmenter} and MaskFormer~\cite{cheng2021per} tackle semantic segmentation from the view of mask classification.
K-Net~\cite{zhang2021k} generates segmentation masks by a group of learnable kernels.
Inspired by the similarity between mask transformers and clustering algorithms~\cite{lloyd1982least}, clustering-based mask transformers are proposed to segment images~\cite{yu2022cmt,xu2022groupvit,yu2022k}.
Deformable transformer~\cite{zhu2020deformable} is also used for improving the image segmentation as in Panoptic SegFormer~\cite{li2022panoptic} and Mask2Former~\cite{cheng2022masked}. 

Similar to this work, Region Proxy~\cite{zhang2022semantic} (RegProxy) also incorporates the idea of superpixels into a deep segmentation network by using a CNN decoder to learn the association between each pixel and superpixel. However, RegProxy uses features on the regular pixel grid to represent each superpixel, and on which to apply self-attention. In comparison, we apply a set of learned weights, which correspond to each superpixel, and use cross-attention with the pixel features to compute the pixel-superpixel associations. Our experiments demonstrate that our methodology provides significant performance improvements.

In summary, all of the prior works that apply transformers to segmentation have, in some way, relied on a dense CNN decoder to generate the final dense features, and then combined these features with an attention mechanism to improve performance. Our method, in comparison, uses cross attention to reduce the image into a small set of superpixels, and only applies self-attention in this superpixel space in the decoder. This allows our method to operate on a significantly lower dimensional space (often $32^2\times$ smaller than the image resolution), while utilizing the benefits that come with global self-attention to achieve state-of-the-art performance.
\section{Method}

Our proposed Superpixel Transformer architecture, summarized in Figure~\ref{fig:architecture}, consists of four main components:
\begin{enumerate}
    \item \textit{Pixel Feature Extraction:} A convolutional encoder backbone to generate hypercolumn features. 
    \item \textit{Superpixel Tokenization:} A series of \textit{local} dual-path cross-attentions, between a set of learned queries and pixel features, to generate a set of superpixel features. 
    \item \textit{Superpixel Classification:} A series of multi-head self-attention layers to refine the superpixel features and produce a semantic class for each superpixel.
    \item \textit{Superpixel Association:} Associating the predicted superpixel classes and pixel features to obtain the final dense semantic segmentation.
\end{enumerate}
We detail each component in the following subsections.

\subsection{Pixel Feature Extraction}
Typical convolutional neural networks, such as ResNet~\cite{he2016deep} and ConvNeXt~\cite{liu2022convnet}, are employed as the encoder backbone.
On top of the encoder output, we apply a multi-layer perceptron (MLP), and bilinear resize to the features after stage-1 (stride 2), stage-3 (stride 8), and stage-5 (stride 32).
The multi-scale features are combined with addition to form hypercolumn features~\cite{hariharan2015hypercolumns}.
Each pixel feature is represented by their corresponding hypercolumn features, which are fed to the following Superpixel Tokenization module.

\subsection{Superpixel Tokenization}
Before introducing our proposed Superpixel Tokenization module, we briefly review the previous works on Differentiable SLIC~\cite{achanta2012slic,jampani2018superpixel}, which we modernize with the transformer framework.

\textit{Preliminary: Differentiable SLIC}~\quad
Simple Linear Iterative Clustering (SLIC)~\cite{achanta2012slic} adopts the classical iterative $k$-means algorithm~\cite{lloyd1982least} to generate superpixels by clustering pixels based on their features (\eg, color similarity and location proximity).
Given a set of pixel features $I_p$ and initialized superpixel features $S_i^0$ at iteration $0$, the algorithm iterates between two steps at iteration $t$:
\begin{enumerate}
\item \textit{(Hard) Assignment:} Compute the similarity $Q_{pi}^t$ between each pixel feature $I_p$ and superpixel feature $S_i^t$. Assign each pixel to a \textit{single} superpixel based on its maximum similarity.
\item \textit{Update:} Update the superpixel features $S_i^t$ based on the pixels features assigned to it.
\end{enumerate}
The Superpixel Sampling Networks (SSN)~\cite{jampani2018superpixel} make the whole process differentiable by replacing the hard assignment between each pixel and superpixel with a soft weight:
\begin{align}
Q_{pi}^t =& e^{-\|I_p - S_i^{t-1}\|^2} \label{eq:SSN_association}\\
S_i^t=&\frac{1}{Z_i^t}\sum_{p=1}^n Q_{pi}^t I_p \label{eq:SSN_update},
\end{align}
where $Z_i^t=\sum_p Q_pi^t$ is the normalization constant.
In practice, at $t=0$, the superpixel features are initialized as the mean feature within a set of rectangular patches that are evenly distributed in the image. In order to reduce the computational complexity and to apply a spatial locality constraint, the distance computation $\|I_p - S_i^{t-1}\|^2$ is restricted to a local $3\times 3$ superpixel neighborhood around each pixel, although larger window sizes are possible.

\begin{figure}[t!]
  \begin{center}
  \includegraphics[width=0.6\linewidth]{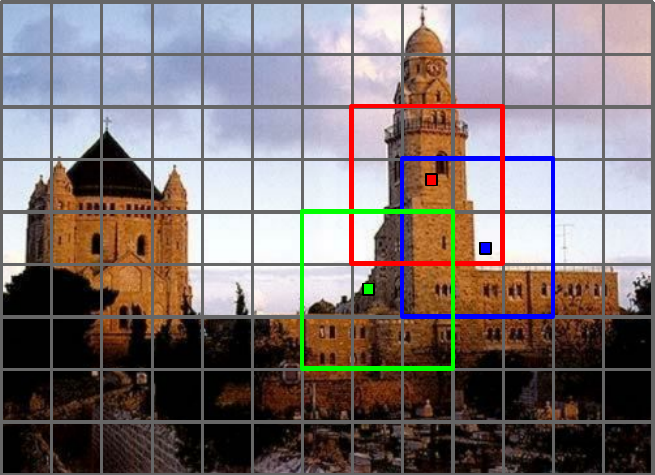} 
  \end{center}
  \caption{Visualization of the superpixel-pixel association. Each superpixel is assigned to a gray grid cell in the image. For each pixel (small dots, size exaggerated) inside a given superpixel, we compute its cross attention with its neighboring 3$\times$3 superpixels, highlighted with the same color. The essence of our method is that these neighborhoods overlap in a sliding window fashion.}
  \label{fig:superpixel_neighborhoods}   
\end{figure}
\textit{Superpixel Tokenization}~\quad
We propose to unroll the SSN iterations and replace the k-means clustering steps with a set of \textit{local} cross-attentions. We initialize the superpixel features, which are distributed on a regular grid in the image, (Figure~\ref{fig:superpixel_neighborhoods}) with a set of randomly-initialized, learnable queries, $S_i^0$, and perform the superpixel update step using cross-attention between superpixel features and pixel features by adapting the dual-path cross-attention~\cite{wang2021max}, giving
\begin{align}
S_i^t=& S_{i}^{t-1} + \sum_{p\in\mathcal{N}(i)}\text{softmax}_p(q_{S_i^{t-1}}\cdot k_{I_p^{t-1}})v_{I_p^{t-1}}\\
I_p^t=& I_p^{t-1} + \sum_{i\in\mathcal{N}(p)}\text{softmax}_i(q_{I_p^{t-1}}\cdot k_{S_i^{t-1}})v_{S_i^{t-1}},
\end{align}
where $\mathcal{N}(x)$ denotes the neighborhood of $x$ and $q$, $k$ and $v$ are the query, key and value, generated applying a MLP to each respective feature plus an additive learned position embedding. For each superpixel neighborhood corresponding to a superpixel, we share the same set of position embeddings. For each superpixel $S_i$, there are $9\cdot h\cdot w$ %
pixel neighbors, where $[h, w]$ is the size of the patch covered by one superpixel, while each pixel has 9~superpixel neighbors. We illustrate this neighborhood in Figure~\ref{fig:superpixel_neighborhoods}.
The local dual-path cross-attention repeats $n$ times to generate the output superpixel, $S_i^{t_n}$ and pixel, $I_p^{t_n}$, features.

This \textit{local} dual-path cross-attention serves three purposes:
\begin{itemize}
    \item Reduce complexity compared to a full cross-attention.
    \item Stabilize training, as the final softmax is only between 9 superpixel features or $9\cdot h\cdot w$ pixel features.
    \item Encourage spatial locality of the superpixels, forcing them to focus on a coherent, local over-segmentation.
\end{itemize}

\subsection{Superpixel Classification}
Given the updated superpixel features from the Superpixel Tokenization module, we directly predict a class for each superpixel using a series of self-attentions. In particular, we apply $k$ multi-head self-attention (MHSA) layers~\cite{vaswani2017attention} to learn global context information between superpixels, producing outputs $F_i$. Performing MHSA on the superpixel features is significantly more efficient than on the pixel features, since the number of superpixels is much smaller. In our experiments, we typically use a superpixel resolution that is \textit{32$^2\times$ smaller} than the input resolution. Finally, we apply a linear layer as a classifier, \textit{producing a semantic class prediction for each refined superpixel feature}, $C_i$. As opposed to the CNN pixel decoders used in other approaches~\cite{cheng2019panoptic,cheng2021per}, our superpixel class predictions $C_i$ can be directly projected back to the final pixel-level semantic segmentation output \textit{without any additional layers}, as described in Section~\ref{sec:superpixel_association}.

\subsection{Superpixel Association}
\label{sec:superpixel_association}
To project the superpixel class predictions back into the pixel space, we use the outputs of the Superpixel Tokenization module, $I_p^{t_n}$ and $S_i^{t_n}$, to compute the association between each pixel and its 9 neighboring superpixels:
\begin{align}
\label{eq:superpixel_association}
Q_{pi}=&\text{softmax}_{i\in\mathcal{N}(p)} (I_p^{t_n}\cdot S_i^{t_n}).
\end{align}
The final dense semantic segmentation, $Y$, is then computed at each pixel, $p$, as the combination of each predicted superpixel class from the Superpixel Classification module, $C_i$, weighted by the above associations:
\begin{align}
Y_p =& \sum_{i\in \mathcal{N}(p)}Q_{pi}\cdot C_i.
\end{align}
During training, the dense semantic segmentation $Y$ is supervised by the semantic segmentation ground truth. %

\begin{table*} [t]
\centering
\begin{tabular}{l|c|ccc|c}
Method & Backbone & Params $\downarrow$ & FLOPs $\downarrow$ & FPS$ \uparrow$ & mIoU $\uparrow$ \\
\shline
MaskFormer~\cite{cheng2021per} & ResNet-50~\cite{he2016deep} & - & - & - & 78.5 \\
Mask2Former\cite{cheng2022masked} & ResNet-50~\cite{he2016deep} & - & - & - & 79.4 \\
Panoptic-DeepLab~\cite{cheng2019panoptic} & ResNet-50~\cite{he2016deep} & 43M & 517G & - & 78.7\\
RegProxy$^*$~\cite{zhang2022semantic} & ViT-S~\cite{dosovitskiy2020image} & \textbf{23M} & 270G & - & 79.8 \\
$k$MaX-DeepLab$^{\dagger}$~\cite{yu2022k} & ResNet-50~\cite{he2016deep} & 56M & 434G & 9.0 & 79.7 \\
\hline
SP-Transformer & ResNet-50~\cite{he2016deep} & 29M & \textbf{253G} & \textbf{15.3} & \textbf{80.4} \\
\shline 
Mask2Former$^{\ddagger}$~\cite{cheng2022masked} & Swin-L~\cite{liu2021swin} & - & - & - & 83.3 \\
RegProxy$^*$~\cite{zhang2022semantic} & ViT-L/16~\cite{dosovitskiy2020image} & 307M & - & - & 81.4 \\
SegFormer~\cite{xie2021segformer} & MiT-B5~\cite{xie2021segformer} & \textbf{85M} & \textbf{1,448G} & 2.5 & 82.4 \\
$k$MaX-DeepLab$^{\dagger}$~\cite{yu2022k} & ConvNeXt-L~\cite{liu2022convnet} & 232M & 1,673G & 3.1 & \textbf{83.5} \\
\hline
SP-Transformer & ConvNeXt-L~\cite{liu2022convnet} & 202M & 1,557G & \textbf{3.6} & 83.1 \\
\hline

\end{tabular}
\caption{Cityscapes \textit{val} set results. We evaluate FLOPs and FPS with input $1024\times2048$ for our SP-Transformer on a Tesla V100-SXM2 GPU. SP-Transformer with ResNet-50 outperforms prior arts in terms of parameters, latency, and performance. For the large models, SP-Transformer with ConvNeXt-L backbone achieves similar reductions in parameters, while achieving the lowest latency, and competitive mIoU performance. $^*$ RegProxy evaluates using a 768$^2$ sliding window. $^\dagger$\textit{k}MaX-DeepLab is trained for panoptic segmentation.}
\label{tab:cityscapes}
\end{table*}
\begin{table*} [t]
\centering
\begin{tabular}{l|c|c|ccc|c}
Method & Backbone & Crop & Params $\downarrow$ & FLOPs $\downarrow$ & FPS $\uparrow$ & mIoU $\uparrow$ \\
\shline
RegProxy~\cite{zhang2022semantic} & ViT-Ti/16~\cite{dosovitskiy2020image} & 512 & \textbf{6M} & \textbf{3.9G} & 38.9 & 42.1\\
MaskFormer~\cite{cheng2021per} &  ResNet-50~\cite{he2016deep} & 512 & 41M & 53G & 24.5 & 44.5 \\
$k$MaX-DeepLab$^\dagger$~\cite{yu2022k} & ResNet-50~\cite{he2016deep} & 641 & 57M & 75G & 38.7 & \textbf{45.0} \\
\hline
SP-Transformer  & ResNet-50~\cite{he2016deep} & 640 & 29M & 78G & \textbf{40.8} & 43.7 \\ 
\hline
\end{tabular}
\caption{ADE20K \textit{val} set results. We evaluate FLOPs and FPS with input $640\times640$ for SP-Transformer on a Tesla V100-SXM2 GPU. Our method outperforms the prior RegProxy work, while remaining competitive with other prior works, and at the highest FPS. $^\dagger$\textit{k}MaX-DeepLab is trained for panoptic segmentation.} 
\label{tab:ade20k}
\end{table*}
\begin{figure*}[t!]
  \begin{center}
  \includegraphics[width=0.3\linewidth]{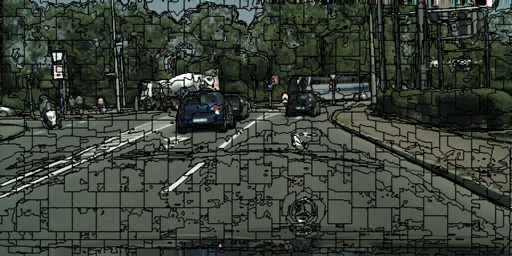}
  \includegraphics[width=0.3\linewidth]{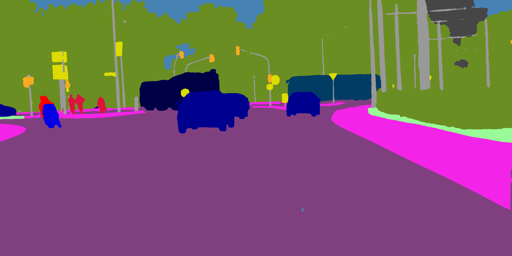}
  \includegraphics[width=0.3\linewidth]{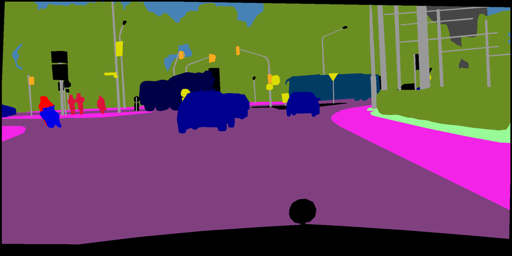}
  
  \includegraphics[width=0.3\linewidth]{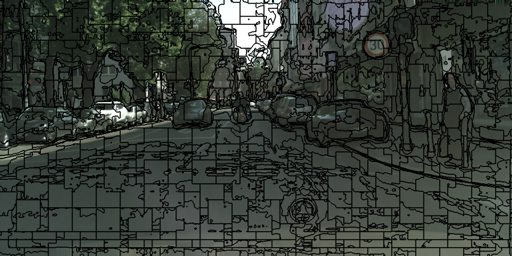}
  \includegraphics[width=0.3\linewidth]{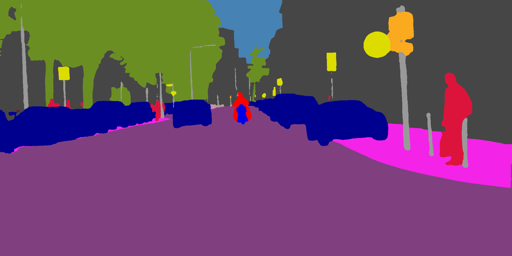}
  \includegraphics[width=0.3\linewidth]{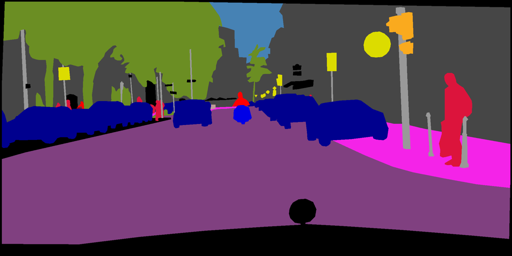}

  \includegraphics[width=0.3\linewidth]{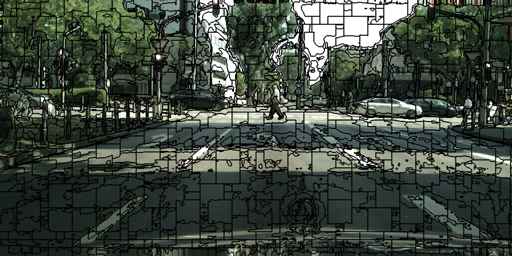}
  \includegraphics[width=0.3\linewidth]{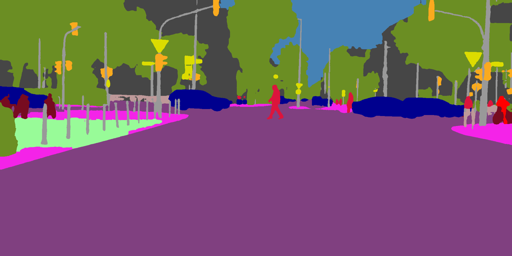}
  \includegraphics[width=0.3\linewidth]{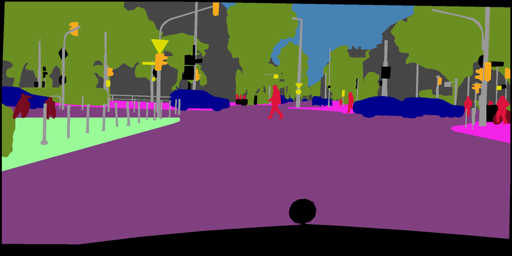}
  
  \vspace{10pt}
  
  \includegraphics[width=0.23\linewidth]{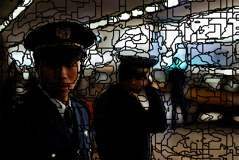}
  \includegraphics[width=0.23\linewidth]{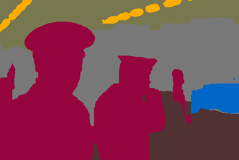}
  \includegraphics[width=0.23\linewidth]{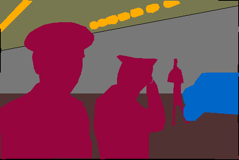}

  \includegraphics[width=0.23\linewidth]{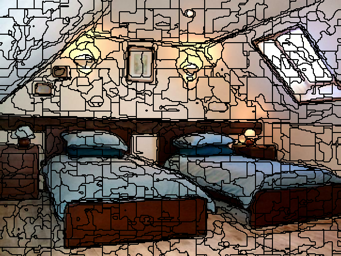}
  \includegraphics[width=0.23\linewidth]{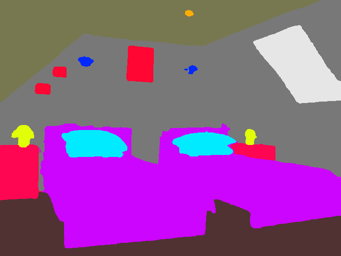}
  \includegraphics[width=0.23\linewidth]{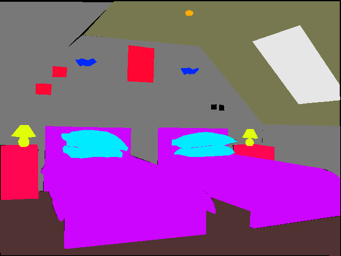}

  \includegraphics[width=0.23\linewidth]{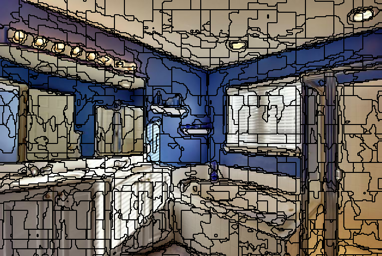}
  \includegraphics[width=0.23\linewidth]{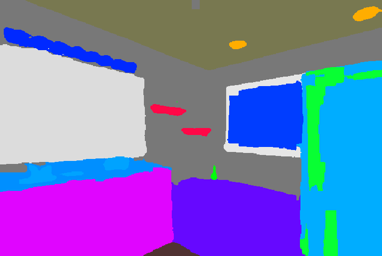}
  \includegraphics[width=0.23\linewidth]{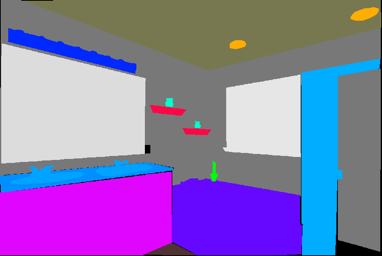}
  
  \end{center}
  \caption{Qualitative examples of our ConvNeXt-L backbone model on the CityScapes (top) and ADE20k (bottom) datasets. Left: Input image with superpixel boundaries overlaid. Middle: Semantic prediction. Right: Semantic ground truth. The superpixels are visualized by generating a hard assignment by taking the argmax of the soft assignment. Despite only being trained with a semantic segmentation loss and with soft assignments, the hard assignment superpixels faithfully follow boundaries in the image. The superpixel overlays are best viewed zoomed in.}  
  \label{fig:cityscapes_qualitative}   
\end{figure*}
\section{Experiments}
In this section, we evaluate the size, latency and accuracy of a small (ResNet-50 backbone) and large (ConvNeXt-L backbone) variant of our model against prior works, and provide ablations for the superpixel tokenization module and a fine grained latency analysis.

We evaluate our work on the Cityscapes~\cite{cordts2016cityscapes} and ADE20K~\cite{zhou2017scene}. Cityscapes is a driving dataset, consisting of 5,000 high resolution street-view images, and 19 semantic classes. ADE20K is a general scene parsing dataset, consisting of 20,210 images with 150 semantic classes.

\subsection{Implementation Details}
\textit{Pixel Feature Extraction}
The hypercolumn features are generated by applying a MLP to project each encoder feature to 256 channels, and bilinear resizing to stride 8.

\textit{Superpixel Tokenization}
For both datasets, we apply 2 sequential local dual-path cross-attention to generate the superpixel embeddings, each with 256 channels and 2 heads. We use a single set of learned position embeddings for each superpixel and pixel feature, initialized at $\frac{1}{4}$ resolution of each feature, bilinear upsampled to the feature resolution and added to the feature.

\textit{Superpixel Classification} 4 multi-head attention layers are applied in the superpixel classification stage, with 4 heads each, outputting 256 channels in each layer.

\textit{Superpixel Association}
The associations between the superpixel features and pixel features are computed at stride 8. The association is then bilinear upsampled to the input resolution before applying the softmax in ~\eqref{eq:superpixel_association}.

\textit{Training}
Our training hyperparameters closely follow prior work such as~\cite{chen2018deeplabv2,yu2022k}.
Specifically, we employ the polynomial learning rate scheduler, and the backbone learning rate multiplier is set to 0.1.
The AdamW optimizer~\cite{kingma2014adam,loshchilov2017decoupled} is used with weight decay $0.05$. %
For regularization and augmentations, we use random scaling, color jittering~\cite{cubuk2018autoaugment}, and drop path~\cite{huang2016deep} with a keep probability of 0.8 (for ConvNeXt-L). We use a softmax cross-entropy loss, applied to the top 20\% of pixels of the dense segmentation output.

\textit{Cityscapes}
Our models are trained with global batch size 32 over 32 TPU cores for 60k iterations.
The initial learning rate is $10^{-3}$ with 5,000 steps of linear warmup.
The model accepts the full resolution $1024\times 2048$ images as input, and produces $128\times 256$ hypercolumn features (\ie, stride 8).
Taking as input the hypercolumn features, the superpixel tokenization module uses $32\times 64$ superpixels.

\textit{ADE20K}
Our models are trained with global batch size 64 over 32 TPU cores and crop size $640\times 640$. Our Resnet-50 and ConvNeX-L models are trained for 100k and 150k iterations, respectively. The initial learning rate is $10^{-3}$ with 5,000 steps of linear warmup. $160\times 160$ hypercolumn features are generated, and $40\times 40$ superpixels are used.

\subsection{Results}
\subsubsection{Cityscapes Dataset}
Table~\ref{tab:cityscapes} compares the results of our Superpixel Transformer model to other transformer-based state-of-the-art models for semantic segmentation on the Cityscapes \textit{val} set. In these experiments, we compare models with backbones roughly the same size as ResNet-50 and ConvNeXt-L. In addition to other semantic segmentation models, we also compare against the state of the art panoptic segmentation method, kMaX-DeepLab~\cite{yu2022k}. While this method is trained on a slightly different task, we find that, for computational cost, it is the most fair comparison, as our training schedule and pipeline is most similar to theirs.

With the smaller ResNet-50 backbone, our model is roughly half the size and latency of kMaX-DeepLab, while improving upon mIoU by 0.6 against the previous SOTA, RegProxy~\cite{zhang2022semantic}. As the ResNet-50 backbone is relatively small, most existing models are dominated by the size of their decoders, which allows our model's reduction in decoder size to have significant impacts on the overall size and performance, with a $70\%$ improvement in FPS compared to kMaX-DeepLab. In addition, we expect to see this effect even more for even smaller backbones.

For comparison, we also provided results with a larger ConvNeXt-L backbone. Here, our model has a similar absolute reduction in params and FLOPs, as compared to the equivalent kMaX-DeepLab model. However, as the model is largely dominated by the size of the backbone, the overall improvements are more modest. Nonetheless, our model is able to achieve near state-of-the-art performance, especially compared to the prior semantic segmentation methods, where we outperform most of the prior works, except for Mask2Former, where we are within 0.2mIoU. We note that, for this comparison, the equivalent Mask2Former~\cite{cheng2021mask2former} model is pre-trained with a significantly larger ImageNet-22k dataset, whereas our model is pre-trained on ImageNet-1k.

Figure~\ref{fig:cityscapes_qualitative} provides qualitative examples of our ConvNeXt-L model. The semantic segmentation predictions suggest that the model is able recover thin structures, such as poles, despite largely operating in a 32$\times$64 superpixel space. 

We also provide a visualization of the learned superpixels. We convert the soft association in Section~\ref{sec:superpixel_association} to a hard assignment by selecting the argmax over superpixels, $i$:
\begin{align}
\bar{Q}_{p}=&\text{argmax}_{i\in\mathcal{N}(p)}Q_{pi}
\end{align}
We visualize the boundaries of these assignments overlaid on top of the input image in the left-most column of Figure~\ref{fig:cityscapes_qualitative}. 

From these visualizations, we can see that, despite the model being trained with a soft association, the superpixels generated by the hard assignment tightly follow the boundaries in the image. We note that this is particularly interesting as we do not provide any direct supervision to the superpixel associations, and instead these are learned implicitly by the network. In addition, we find that these boundaries tend to be more faithful to the edges of an object than the labels.

\begin{figure}[t!]
  \begin{center}
  \includegraphics[width=0.329\linewidth]{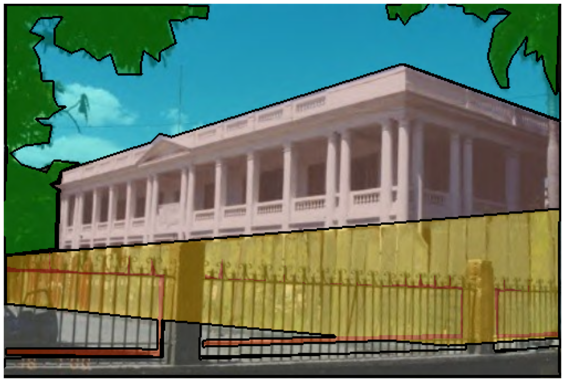}
  \includegraphics[width=0.287\linewidth]{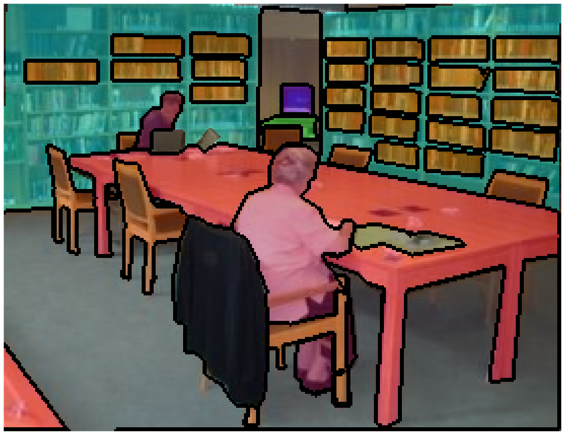}
  \includegraphics[width=0.385\linewidth]{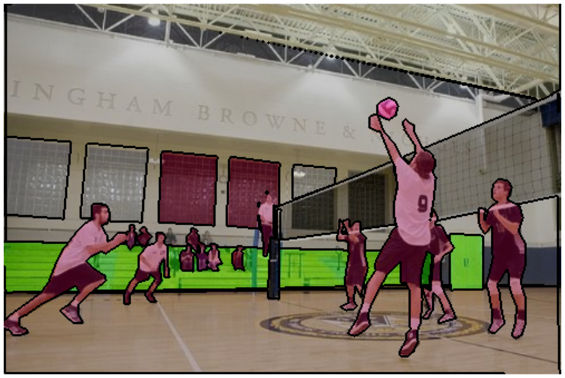}
  \includegraphics[width=0.193\linewidth]{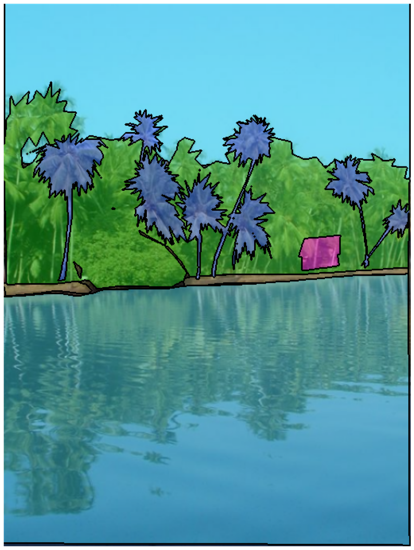}
  \end{center}
  \caption{Inconsistent label boundaries in the ADE20K dataset make it difficult for our model to effectively learn superpixels. See the spaces in between fence railings, the books on the shelves, the unlabeled people on the bleachers and the trees/vegetation.}  
  \label{fig:ade_issues}
\end{figure}
\subsubsection{ADE20K Dataset}
We provide quantitative results on the ADE20K dataset in Table~\ref{tab:ade20k}. We choose one of the most commonly used crop sizes (640$\times$640) and provide results for the ResNet-50 backbone. 

For ADE20K, we achieve the highest FPS and the second lowest \# params (behind the surprisingly small RegProxy model), while outperforming RegProxy.

However, we do note that the gap in performance is larger for ADE20K than Cityscapes. Our hypothesis is that the large number of classes in ADE20K (152) results in ambiguities when a pixel could belong to multiple classes. This results in inconsistencies for object classes, and in particular where boundaries are drawn (see Figure~\ref{fig:ade_issues} for examples). As our superpixel tokenization module operates before any semantic prediction, and each superpixel query only operates on a local neighborhood in the pixel space, the model must learn a consistent way to divide the image into a set of superpixels. When the label boundaries are inconsistent, our model is less able to effectively learn this over-segmentation, leading to a small decrease in performance.

\begin{table} [t]
\centering
\begin{tabular}{cccccc}
\# CA & sp. res. & params & FLOPs & Runtime(ms) & mIoU \\
\shline
1 & 32$\times$64 & 28M & 234G & 50.0 & 78.0 \\
2 & 16$\times$32 & 29M & 240G & 53.5 & 74.9 \\
2 & 32$\times$64 & 29M & 253G & 63.4 & 80.4 \\
2 & 64$\times$128 & 31M & 404G & 120.5 & 79.7
\end{tabular}
\caption{Experiments ablating  the number of local cross-attention layers (\# CA) and the resolution of the superpixels. Ablations are performed with a ResNet-50 backbone on Cityscapes.}
\label{tab:tokenization_ablation}
\end{table}
\subsubsection{Superpixel Tokenization Ablation}
In Table~\ref{tab:tokenization_ablation}, we provide an ablation of the number of cross attention layers and the superpixel resolution, evaluated using our ResNet-50 backbone model on Cityscapes. 

From these experiments, we find that increasing the number of cross attention layers from 1 to 2 improves mIoU significantly by 2.4. Further increasing the number of these layers may have further improvements in performance, although we are currently bottlenecked by accelerator memory constraints. This is largely due to our naive TensorFlow implementation~\cite{weber2021deeplab2}, which we discuss in Section~\ref{sec:latency}.

Reducing the number of superpixels to 16$\times$32 also has a significant regression in mIoU of 5.5. On the other hand, increasing the superpixels to 64$\times$128 also results in a mild regression of 0.7. We posit that this is because our pixel features used in the association are at stride 8 (128$\times$256). As a result, having 64$\times$128 superpixels provides each local-cross attention with a neighborhood of $128/64\times 3$ = 6$\times$6 pixels, which makes the receptive field for each superpixel too small to learn the oversegmentation as effectively. This can be resolved by increasing size of the neighborhood around each superpixel, but at significantly higher computational cost.

\subsection{Discussion}
\subsubsection{Superpixel Quality}
Despite our network not being trained with any explicit superpixel-based loss, we find that the associations learned by the network closely resemble classical superpixels. That is, the superpixels are aligned such that they follow the dominant edges in the image. We posit that this is due to the limited receptive field for each superpixel's cross attention. As any single superpixel may not have visibility to all of the pixels for a given mask, the model must use local edges and boundaries to separate the superpixels.

\subsubsection{Latency} \label{sec:latency}
As seen in Tables~\ref{tab:cityscapes} and ~\ref{tab:ade20k}, our method provides a significant improvement in FPS. As the main processing of our model operates on the small superpixel space, this allows for a large reduction in model complexity, while achieving state of the art performance.

However, we believe that there is still a large further reduction in latency available. In particular, the local cross attention operation is not efficient for standard accelerators and native TensorFlow or PyTorch implementations. This is because it requires a sliding window with overlapping patches, but with different operands in each patch (as opposed to convolutions where the weights are the same for all patches). However, as we operate on the superpixel grid level (32$\times$64 for Cityscapes and 40$\times$40 for ADE20K), this impact of this inefficiency is low enough to make our method overall faster compared to methods which operate on the dense pixel space.

Nonetheless, the Superpixel Tokenization module takes up the majority of the decoder runtime, as can be seen in Table~\ref{tab:cityscapes_timing}, where we provide timing information for our method with a ResNet-50 backbone on a 1024$\times$2048 input. Our experiments are currently designed with a relatively naive, pure TensorFlow implementation, and involves the duplication of each superpixel or pixel 9 times (depending on the cross-attention operation). We believe that a CUDA implementation could remove this redundant copy, and provide even further speedups for our method.

\begin{table} [t]
\centering
\begin{tabular}{l|c}
Method & Time (ms) \\
\shline
Backbone: ResNet-50 & 28.0\\ %
Hypercolumn & 7.1\\
Superpixel Tokenization & 25.4 \\ %
Superpixel Self-Attention & 4.0\\
Superpixel Association & 1.0\\
\hline
Total & 65.5 \\
\shline

\end{tabular}
\caption{Latency information for each sub-component. The bulk of the non-backbone latency is consumed by the local cross attention in the superpixel tokenization module.}
\label{tab:cityscapes_timing}
\end{table}
\section{Conclusions}
We presented a novel network architecture for semantic segmentation that leverages superpixels to project the dense image segmentation problem into a low dimensional superpixel space. Operating on this space enables us to significantly reduce the size and inference latency of our network compared to prior works, while achieving state-of-the-art performance.

{\small
\bibliographystyle{ieee_fullname}
\bibliography{egbib}
}

\end{document}